\documentclass[conference]{IEEEtran}
\IEEEoverridecommandlockouts
\usepackage{cite}
\usepackage{amsmath,amssymb,amsfonts}
\usepackage{algorithmic}
\usepackage{graphicx}
\usepackage{textcomp}
\usepackage{xcolor}
\def\BibTeX{{\rm B\kern-.05em{\sc i\kern-.025em b}\kern-.08em
    T\kern-.1667em\lower.7ex\hbox{E}\kern-.125emX}}

\newcommand{\mnet}[0]{\textsc{Motion-Net}}
\newcommand{\mnets}[0]{\textsc{Motion-Nets}}

\begin{document}

\title{\mnets: 6D Tracking of Unknown Objects in Unseen Environments using RGB\\
% \thanks{TODO: Identify applicable funding agency here. If none, delete this.}
}

\author{\IEEEauthorblockN{Felix Leeb}
\IEEEauthorblockA{\textit{Computer Science and Engineering} \\
\textit{University of Washington}\\
Seattle, USA \\
fleeb@cs.washington.edu}
\and
\IEEEauthorblockN{Arunkumar Byravan}
\IEEEauthorblockA{\textit{Computer Science and Engineering} \\
\textit{University of Washington}\\
Seattle, USA \\
barun@cs.washington.edu}
\and
\IEEEauthorblockN{Dieter Fox}
\IEEEauthorblockA{\textit{Computer Science and Engineering} \\
\textit{University of Washington}\\
Seattle, USA \\
fox@cs.washington.edu}

}

\maketitle

\begin{abstract} 
In this work, we bridge the gap between recent pose estimation and tracking work to develop a powerful method for robots to track objects in their surroundings.~\mnets~use a segmentation model to segment the scene, and separate translation and rotation models to identify the relative 6D motion of an object between two consecutive frames. We train our method with generated data of floating objects, and then test on several prediction tasks, including one with a real PR2 robot, and a toy control task with a simulated PR2 robot never seen during training.~\mnets~are able to track the pose of objects with some quantitative accuracy for about 30-60 frames including occlusions and distractors. Additionally, the single step prediction errors remain low even after 100 frames. We also investigate an iterative correction procedure to improve performance for control tasks.
\end{abstract}


\vspace{-2mm}
\section{Introduction}
%While even toddlers can identify and manipulate objects they have never seen before with ease, the same cannot be said of robots. Much of the effort of modern robotics goes into developing learning algorithms that allow a robot to perceive and manipulate arbitrary objects around it in dynamic or new environments. With the onset of deep convolutional neural networks, great advances have been made in computer vision. However for many robotics tasks the traditional machine learning approach of training a monolithic model from a large carefully labelled dataset, is not sufficient because (1) collecting datasets with real robots risks damaging the robots, (2) labelling the data is very time-intensive, (3) the resulting models usually generalize poorly to new settings. Meanwhile, collecting data using a simulation with domain-randomization [TODO:ref] alleviates all these issues.

% more motivation

% For a robot to perform complex tasks and engage with humans, it is necessary for the robot to perceive objects it has potentially never seen before. With the onset of deep convolutional neural networks, great advances have been made in computer vision. However for many robotics tasks the traditional machine learning approach of training a monolithic model from a large carefully labelled dataset, is not sufficient because (1) collecting datasets with real robots risks damaging the robots, (2) labelling the data is very time-intensive, (3) the resulting models usually generalize poorly to new settings. 

Humans are able to track arbitrary objects moving through space with relative ease (even without depth information, such as in a video). While it is perhaps infeasible for humans to be able to estimate the 3D translation and 3D rotation of objects with quantitative accuracy, for many common tasks tracking the qualitative motion of objects, such as the curve of a spinning soccer ball, is sufficient to react.

Much of the effort of modern robotics goes into developing learning algorithms that allow a robot to perceive and manipulate arbitrary objects around it in dynamic or new environments. The traditional machine learning approach of training a monolithic model from a large carefully labeled dataset is not sufficient for such tasks, as it is often infeasible to collect a sufficiently large dataset, especially when dealing with real robots. Instead, we aim to give our robot some intuition over how rigid objects move independent of the specific shape, color, or background using transfer learning~\cite{tan2018survey} with domain randomization~\cite{tobin2018domain,tremblay2018training}, which provides us with the necessary training data, while still allowing our model to perform in real world settings.

There is already a wealth of literature on tracking unknown objects in unseen environments~\cite{gordon2018re,alvar2018mv}, however these usually only track objects using bounding boxes, which is insufficient for many robotics tasks, such as grasping, where the full 6D pose (3D position and 3D orientation) is usually assumed to be known. Meanwhile, most frameworks in pose estimation require some information about the object such as the object class, a 3D model of the object~\cite{li2018deepim,tekin2018real,xiang2017posecnn}, or are trained on a specific objects~\cite{byravan2017se3}. To meld these two perspectives, we present~\mnets~which estimates the relative 6D motion of objects over time for both prediction and control tasks. Our primary contributions herein are:

\begin{enumerate}
    \item We combine ideas from recent work on pose estimation and tracking to produce a more general method for 6D object tracking.
    \item We employ transfer learning for both simulated and real prediction tasks, and several simulated control tasks to test our model.
    \item We develop an iterative correction procedure to allow our model to correct accumulating errors from integrating single step open-loop motion estimates.
\end{enumerate}

% emphasize open-ended learning and transfer learning - mnets are a sort of stepping stone, get a first approximation for any object in any environment, then some other method can work on categorizing the object and further learning about it.

% contributions
% - combine ideas from pose estimation and tracking to produce a more general method for 6D object tracking
% - investigate transfer learning for prediction and control by training in simulation and testing in various simulated and real settings
% - develop an iterative correction procedure to allow our model to correct the open-loop long term motion prediction

% include prior work here (?)

\section{\mnets}

\mnets~split the tracking problem into two components: (1) segmentation and (2) motion estimation. First, the Segmentation model (described in \ref{sec:seg}) identifies the object being tracked from the background, generating a segmentation mask of the target object. Then, a Translation and Rotation model (described in \ref{sec:motion}) estimate the observed translation and rotation between two consecutive frames, respectively. Aside from the RGB images at each timestep, the only input to the network is a single dense segmentation mask for the first frame, which is akin to telling the network which object should be tracked (similar to assumptions made by bounding box trackers~\cite{gordon2018re}).
%  These single step motions are then integrated in time to track the object over multiple frames.

\subsection{Segmentation} \label{sec:seg}

\begin{figure}
    \begin{center}
        \includegraphics[width=.4\textwidth]{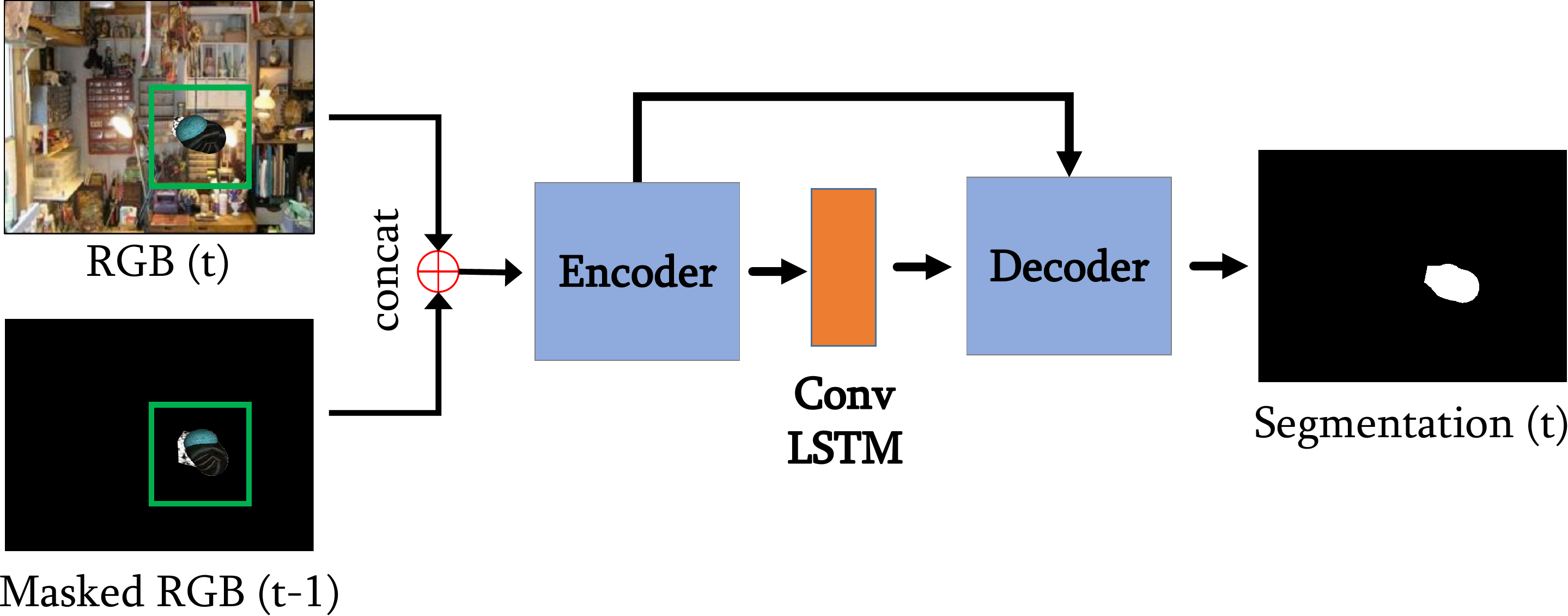}
        \caption[Segmentation Model Overview]{A high level overview of how the segmentation model produces a mask of the object in the current frame, using the masked out object in the previous frame. From the full RGBs only the crops (in green) are concatenated and passed through the model.}
        \label{fig:seg}
    \end{center}
\end{figure}

The segmentation model (seen in figure~\ref{fig:seg}) takes as input the previous RGB, previous mask, and current RGB frame. The output of the segmentation model is a dense segmentation mask of the object for the current timestep.

First, the previous mask is used to crop out the object in the previous and current RGB. Since the object is presumably not in the exact same position from one frame to the next, the crop is scaled to 40\% of the image height with an aspect ratio set to 1:1, so there is some of the background is included around the object. This cropping allows the segmentation model to focus on a relatively small part of the scene without losing the object if it moves around, similar to the work in~\cite{gordon2018re}. To further focus the model, the object in the previous frame is masked out using the previous mask. The masked previous RGB and the current RGB are then concatenated along the channels dimension and passed into an encoder with seven convolutional layers, a single 128 channel Conv-LSTM layer~\cite{xingjian2015convolutional}, and finally a seven layer decoder mirroring the encoder with skip-add connections. Overall the segmentation model has approximately 1.5M trainable parameters.
%First, the previous mask is used to compute the crop parameters of the current image. The center of the crop is given by the center of the object mask, and the width and height are chosen such that the crop height is 40\% of the original image height and the aspect ratio is 1. These crop parameters are then used to crop all inputs (both RGBs and the previous mask). Next, the crops are concatenated along the channels, there are a total of seven channels (three from each of the RGBs and one from the previous mask), and then resampled using bilinear upsampling to 128x128. This 7x128x128 input image is then passed through a deep convolutional network using an encoder-decoder architecture with skip-add connections, for a total of ten convolutional layers with instance normalization and a PReLU nonlinearity after each convolutional layer and a 128 channel Conv-LSTM layer~\cite{xingjian2015convolutional} in the middle of the encoder and decoder. Finally, the 1x128x128 output mask is subsampled using bilinear downsampling to the original crop size, and replaced using the crop parameters. Overall the segmentation model has approximately 1.5 million trainable parameters.

In principle, any model can be used to extract the dense segmentation masks of the object for the full sequence of RGBs. However, this particular model takes advantage of two important concepts, cropping the original image to focus the network on the object of interest, and using a recurrent state to remember features of the object across the sequence. Since the deep network only ever processes the upsampled crops, the crop size (in this case 40\% of the original height) should be chosen to be as small as possible so the segmentation model doesn't get confused by other objects or the background, all the while still being sufficiently large such that the full object will be visible in both RGB frames. Meanwhile, the recurrent state in the form of the Conv-LSTM hidden layer and cell states allow the model to keep track of visual features of the object such as color, size, and shape, as well as form a motion prior for later frames in the sequence.

\subsection{Motion Estimation} \label{sec:motion}

\begin{figure}
    \vspace{-1mm}
    \begin{center}
        \includegraphics[width=.4\textwidth]{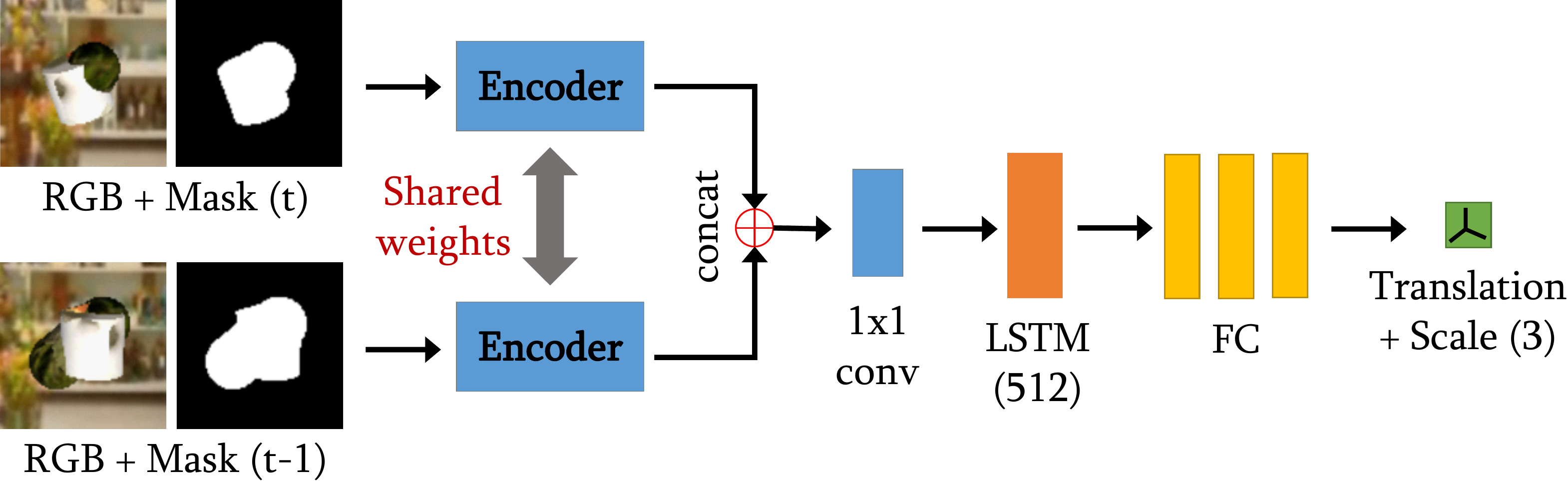}
        \caption[Translation Model Overview]{A high level overview of how the translation model estimates the in plane translation and scale of the object between two consecutive frames using a crop of the RGB at time t and t-1 informed by the mask at time t-1.}
        \label{fig:trans}
    \end{center}
    \vspace{-1mm}
\end{figure}

Both the rotation and translation models use the same basic architecture (as seen in figures~\ref{fig:trans} and~\ref{fig:rot}), with two important differences. First, while both models crop the input images before passing them through a deep convolutional network, the rotation model uses a different cropping method than the translation model which is described below. Secondly, the outputs of the models are processed in different ways to be interpreted as a 3D rotation and 3D translation respectively. Each of the two models has approximately 700k total trainable parameters.

Other than that, both models take the previous RGB and mask as well as the current RGB and mask as input. First, each RGB and corresponding mask are concatenated together with two additional coordinate channels as in~\cite{liu2018intriguing} so there is one 6 channel (RGB + mask + row coordinate + column coordinate) for the current and one for the previous timestep.

After cropping the input images as specified below, each input image is passed through a deep convolutional encoder with six layers to produce one 128x4x4 feature map for the current and one for the previous frame. These two feature maps are then concatenated along the channels and passed through a single 1x1 convolutional layer. This layer is meant to act as a feature correlation layer which compares the features of both frames on a pixel by pixel basis. The resulting features are flattened and then passed through a 512 unit LSTM layer and then a fully-connected network with three hidden layers to produce the output rotation or translation.

\subsubsection{Translations}

We use the non-metric representation of translations as described in~\cite{li2018deepim}. Instead of predicting a motion along the axes of the camera frame, we predict a horizontal and vertical translation in the image space in addition to a scale change (see ~\cite{li2018deepim} for more details). This allows the single step motion predictions to be integrated without any information about the true 6D pose of the object being tracked.

To make it easier for the model to identify the correct translation and scale change, we want to highlight how the location of the object has changed from the previous frame to the current one. This is accomplished by computing the center of the mask of the object in the previous frame, using a fixed height and width with respect to the original image size (here chosen to be 40\% of the height, with a 1:1 aspect ratio). These crop parameters given by the previous frame are applied to both the previous and current input images, so if the object translates in the current timestep, the object will be slightly off center in the crop of the current frame. We choose the relatively large crop size so that even if the object is relatively large or the motion is large, the object is still visible in the current frame using the crop parameters from the previous frame.

\subsubsection{Rotations}

The rotation model is tasked with estimating the 3D rotation of the object between the previous and current frames, in the camera frame. There are several popular ways to represent general SO(3) rotations for deep models including Euler angles, quaternions, axis angle, and the continuous 6D representation proposed by~\cite{zhou2018continuity}. However, we can expect the motion between two consecutive frames to be rather small, so our representation should focus on small rotations. Consequently, we use the gnomonic projection discussed in~\cite{hartley2013rotation}, which neatly avoids the double coverage problem of quaternions by projecting the quaternion sphere in $\mathbb{R}^4$ onto the tangent hyperplane intersecting with $(1, 0, 0, 0)$ (corresponding to $(qw, qx, qy, qz)$ respectively). The only theoretical downside to the gnonomic projection is that $180^{\circ}$ rotations cannot be modelled as they correspond to vectors with infinite magnitude. The gnomonic projection is implemented by having the network predict the $qx$, $qy$, and $qz$ components of a quaternion, while $qw$ is fixed to 1, and then the quaternion is normalized to a versor in the end.

\begin{figure}
    \vspace{-1mm}
    \begin{center}
        \includegraphics[width=.4\textwidth]{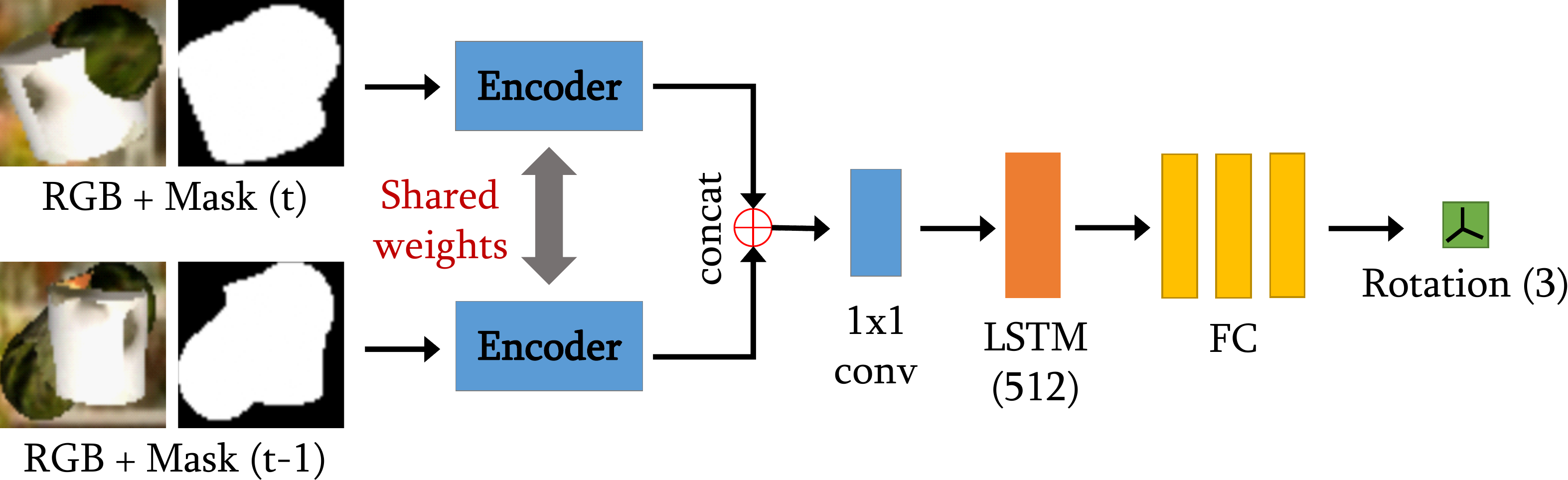}
        \caption[Rotation Model Overview]{A high level overview of how the rotation model estimates the rotations of the object between two consecutive frames using a tight crop of the RGB at time t and t-1 informed by the corresponding masks.}
        \label{fig:rot}
    \end{center}
    \vspace{-1mm}
\end{figure}

Unlike translations, where some background information may help localize the object in the current frame, for rotations we want the network to focus on features on the object and its silhouette. Therefore, we maximize the object in the crop by computing the center and standard deviation from the center of the mask of the object independently for the previous and current frames. The crop is centered on the object mask with a width and height of 2.1 times the standard deviation along the width and height respectively. This way, if the object is particularly tall or wide in the image, then the crop will still predominantly contain the object, rather than the background. A potential disadvantage to this process is that since the height and width of the crop are computed independently, the aspect ratio of the crop may change as the object rotates. We resample the crop using bilinear resampling to fix the aspect ratio at 1:1, which consequently introduces some warping, which can be seen when comparing the example input crops in figures~\ref{fig:trans} and~\ref{fig:rot}.
%(see figure~\ref{fig:TODO} for examples)
However, we find that as long as the subsequent convolutional neural network has a sufficiently large capacity, performance is not negatively impacted by the warping. A significantly bigger issue is that the crop parameters are independently computed for the previous and current frames based on their masks, so the rotation model is significantly more sensitive to masks that change rapidly, such as in the presence of occlusions.

\subsection{Control} \label{sec:ctrl}

Since the~\mnets~directly estimate the motion from frame to frame, the ideal control space would be in the velocity space of the object, as opposed to the position space. For control tasks that require reasoning over positions of objects, the predicted motions must be integrated to keep track of the pose with respect to the original pose of the object. However, when integrating the single timestep motion predictions by the~\mnets~the errors will compound.

For some control problems, the performance loss due to the accumulating errors from integrating the deltas may be mitigated with an online correction procedure, similar to the use of the inverse model in~\cite{agrawal2016learning}. We assume that given a predicted motion in the camera frame between two frames, we have a controller that is able to realize that motion, for example, a robot that can apply forces and torques to the object of interest using its manipulator. Given some target frame where the object of interest has the desired pose, if the object is already sufficiently close to this pose, then we can use the~\mnets~to estimate the pose delta between the target and the current frame, even though they are not necessarily separated by a single timestep worth of motion. While the estimated delta may not be quantitatively accurate, we can expect the motion to be in the right general direction, thereby bringing the object pose closer to the target pose. This suggests an iterative process somewhat like the pose refinement in DeepIM~\cite{li2018deepim}, except instead of rendering the object with the hypothesized pose, since~\mnets~operate without 3D models of the objects, we use a controller to physically move the object and see whether that's closer to the target.

\section{Training and Experiments}
While in principle all three components of~\mnets~(the segmentation model, translation model, and rotation model) could be trained end-to-end, training each model separately is significantly simpler. A specific issue with end-to-end training is that since all models use the masks extensively to compute crop parameters, the very noisy initial predicted masks by the segmentation model will most likely not provide a good training signal to the rotation and translation models. Instead, all models are trained from scratch independently, using the same dataset of simulated random floating objects as specified below. 

\subsection{Data Generation}

All training data is collected with an custom environment of floating objects using the MuJoCo physics engine~\cite{todorov2012mujoco}. We collect two separate training sets, one with a single object per sequence, and one dataset with three moving objects to provide examples of distractor objects and occlusions. Each object is composed of three MuJoCo primitives (sphere, box, cylinder, ellipsoid, capsule) with randomly sampled size parameters making the objects roughly 20-30 cm wide. Each primitive of an object is connected to the others with small random offsets and orientations with respect to each other and is given a texture randomly sampled from the Describable Textures Dataset~\cite{cimpoi2014describing} containing around 5000 textures. The background is sampled from the training set of the Places365 dataset~\cite{zhou2017places} containing around 1.8 million images of indoor and outdoor scenes. All objects begin in a random position and orientation in the scene, but are confined to a box of one cubic meter so they always stay more or less within the view of the camera. The camera is static in all sequences as well as all experiments. In principle, there is no reason~\mnets~can't be trained with moving camera scenes, however, note that~\mnets~will always predict the motion of the objects in the camera frame, so the camera motion must be known to recover the motion of the objects in the world space.

There are about 50k independent sequences in each of the two training sets. Each sequence is 31 frames long and for each timestep a random force and torque is applied to each object, so that each object moves on average 50 cm/s and rotates around 0.5 revolutions/s (at 30 frames/s). In addition to the RGB images, which are rendered with a 240x320 pixel resolution, we also collect the ground truth segmentation mask and pose for each object as supervision.

%  To get an idea for how fast the objects are moving figure~\ref{fig:hist} shows histograms of the rotational and translational motion.
% \begin{figure}
%     \begin{center}
%         \includegraphics[width=.3\textwidth]{figures/valset-motion-histogram.pdf}
%         \caption[Motion Histograms]{Histograms of the motion of the floating objects in the test set (however kinematics are sampled with same distributions in the training set). If the camera is taken to be 25 Hz, then after one second the object has translated on average about half a meter and rotated by almost $120^\circ$}
%         \label{fig:hist}
%     \end{center}
% \end{figure}

In each iteration during training, a single sample comprises a full 31-frame sequence of an object, so the sequences with three objects are used three times, once for each object. Every training iteration uses a batch of 16 samples, where the sequence of each sample in the batch is reversed with a probability of 0.5. A small amount of random noise ($\pm 0.03$) is added to the RGBs after normalizing to 0-1. The recurrent state of each model is reset at the beginning of each iteration, but then updated across the 31 frame sequence. The loss function for the segmentation model is the binary cross entropy function, and the MSE between the prediction and the true non-metric translation for the translation model. The rotation model is trained to minimize the geodesic angle between the predicted and the true rotational motion observed. All models are trained for 14 epochs, using RMSProp with a learning rate of 2e-4 and a weight decay of 1e-6 and a learning rate decay by a factor of 4 every 6 epochs.

During training the translation and rotation models have access to the ground truth segmentation masks, however at test time they use the masks predicted by the segmentation model. We use a curriculum learning scheme so that over the course of training the translation and rotation models become accustomed to using the predicted masks, rather than the ground truth.

\subsection{Prediction}

We evaluated~\mnets~on two prediction tasks: the first is a pair of test datasets very similar to the training sets, except with newly sampled objects and backgrounds. The backgrounds are sampled from the validation set of the Places365 dataset containing about 46K images, so all backgrounds are novel.

In the second prediction task, we tested how well the~\mnets~transfer from tracking simulated floating objects, to real objects moved by a real PR2 robot. Four 100 step motions were recorded at about 31 Hz with a Kinect (only using RGB) camera at a resolution of 480x640 (2x of the training data) for four different objects and evaluated using the same models without any fine tuning, so the~\mnets~have never seen a robot arm during training. As this task is significantly more challenging, the motions were sampled such that there was very little motion in the forward and backward directions, which additionally caused issues due to significant lighting changes on the object during a motion. Since the PR2 robot holds on to the objects throughout the whole motion, we compute the ground truth poses from joint angles using forward (robot) kinematics, technically this only gives us the pose of the gripper, however that is a constant offset from the true object poses.

\subsection{Control}

To demonstrate simple examples of how~\mnets~can be used for control, we evaluate on one small toy setting based on the floating objects, and one using a PR2 robot simulated in MuJoCo. For both settings, an object starts in some randomly sampled pose, then the object is perturbed by some random forces and torques for $T$ steps during which the~\mnets~track the motion of the object, and after that, the task is for the controller to move the object back to it's original pose based on the relative motion estimates from~\mnets. We use the same success criterion as in~\cite{li2018deepim} called the "5cm-$5^\circ$" rule, so the end position and orientation of the object must be within 5cm and $5^\circ$ of the target pose. In this setting, we have a frame with the object in the target pose, so we can use the iterative correction procedure described in~\ref{sec:ctrl}. If the controller has not succeeded after executing the motions from the open-loop predictions, we execute some fixed number of correction iterations to try to reach the 5cm-$5^\circ$ criterion.

Since our focus here is on the tracking by the~\mnets~we make some (rather strong) simplifying assumptions that the controller can apply forces and torques in the camera frame. In the toy settings, this is accomplished by simply applying forces and torques to the object directly. However, for the PR2 task, we use an Inverse Kinematics solver to predict joint positions that will achieve the desired motion of the object, which stays fixed to the robot's end effector. Additionally, we test the simulated PR2 setting using two different views: a first person view of a camera mounted on the robot, and a third person view of a camera looking at the robot.

%Obviously, the task of moving the object to its original pose is trivial when recording the joint positions from the beginning, so a more realistic scenario is to have a controller than can affect the object of interest, however the perturbations are caused by some other process (perhaps a human) so you don't have the joint angles at test time. Even so as long as the controller can apply forces to the object with respect to the camera, it doesn't matter if the controller is just some nameless force in a simulator, or the motions computed by an IK solver, in any case~\mnets~can track the full pose of the object.

\section{Results and Discussion}

\subsection{Prediction}

\subsubsection{Floating-Object}

\begin{figure*}
    \begin{center}
        \includegraphics[width=.8\textwidth]{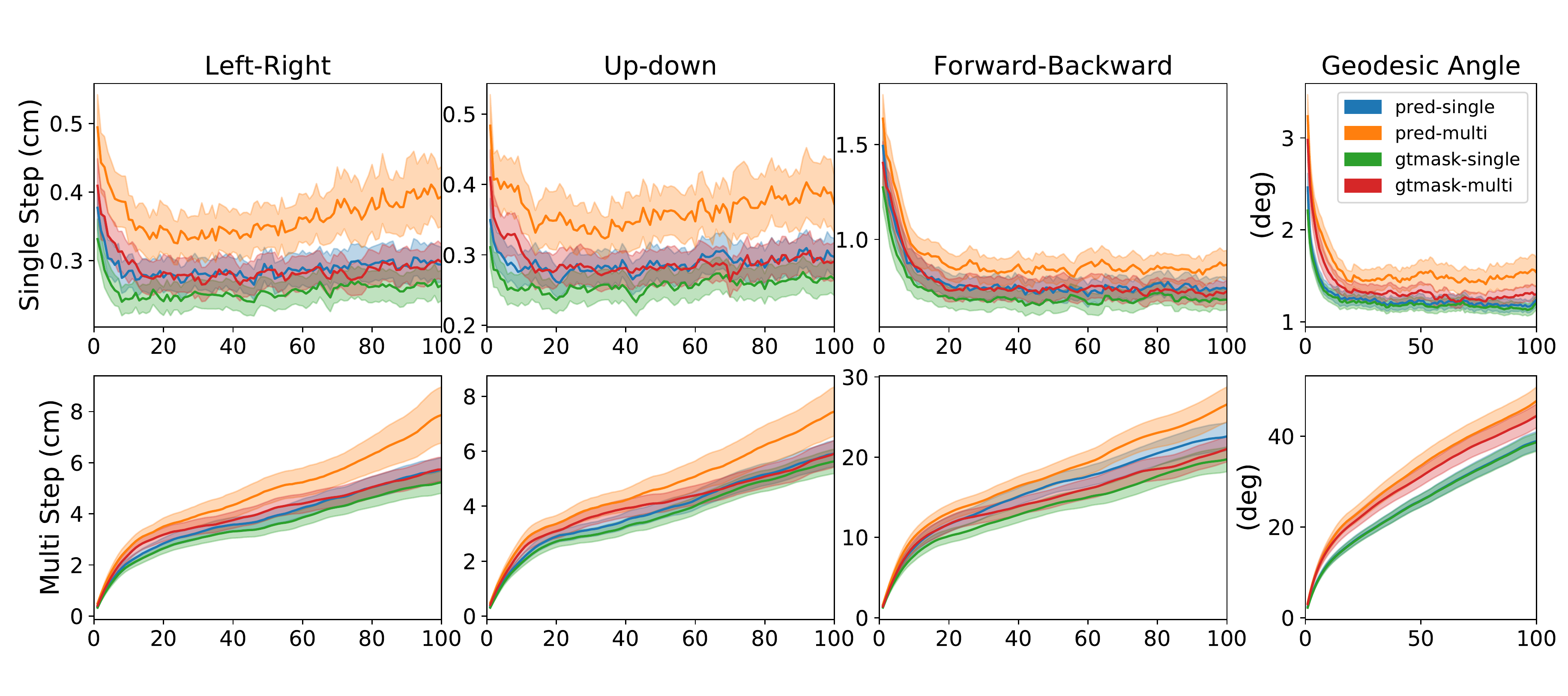}
        \vspace{-5mm}
        \caption[Floating Object Prediction Results]{Errors for single step and integrated (multi step) motion for the floating object setting in an unseen environment with unknown objects. In the key the 'pred' refers to using the masks predicted by the segmentation model, while 'gtmask' means the translation and rotation models are using the ground truth segmentation. 'Multi' refers to the setting with three objects (only one of which is being tracked), so there are occlusions and distractors, while 'single' only has one moving object. The x axis correspods to the frame number in the sequence.} 
        \label{fig:float-pred}
        \vspace{-6mm}
    \end{center}
\end{figure*}

As can be expected, the lack of depth information results in a drastic difference in performance between the in image plane translations (left-right and up-down motion) and the scale changes (forward-backward motion), seen in figure~\ref{fig:float-pred}.

A more encouraging feature is in how the single step errors change as the sequence progresses. For the first ten frames or so the errors per step rapidly decrease, which we can interpret as the model storing information about the object appearance and past motion in the recurrent state. After the recurrent state has been adequately initialized the single step errors remain fairly low, and don't increase significantly overtime. This is very encouraging, as the recurrent states were only trained on sequences of 30 steps, while the test sequences are all 100 steps long, and yet the recurrent state does not drift significantly. There is a slight increase in the single step errors for the model using the predicted masks in the three object environment (pred-multi), however this additional source of error is most likely the segmentation model, which can occasionally get confused by which object it should be tracking, especially in the face of large occlusions early in the sequence. This last problem may be addressed by explicitly modelling the uncertainty of our model in the precise object segmentation, and potentially allowing our model to switch to a different object if it gets confused in the middle of a sequence because the current object differs too much from earlier frames.

Finally, as~\mnets~predict single step deltas in an open loop, integrating these single step predictions leads to increasing errors. This underscores the importance of using some sort of correction procedure as will be discussed in section~\ref{sec:ctrl}.

\subsubsection{Real Robot}

\begin{figure*}
    \begin{center}
        \includegraphics[width=.8\textwidth]{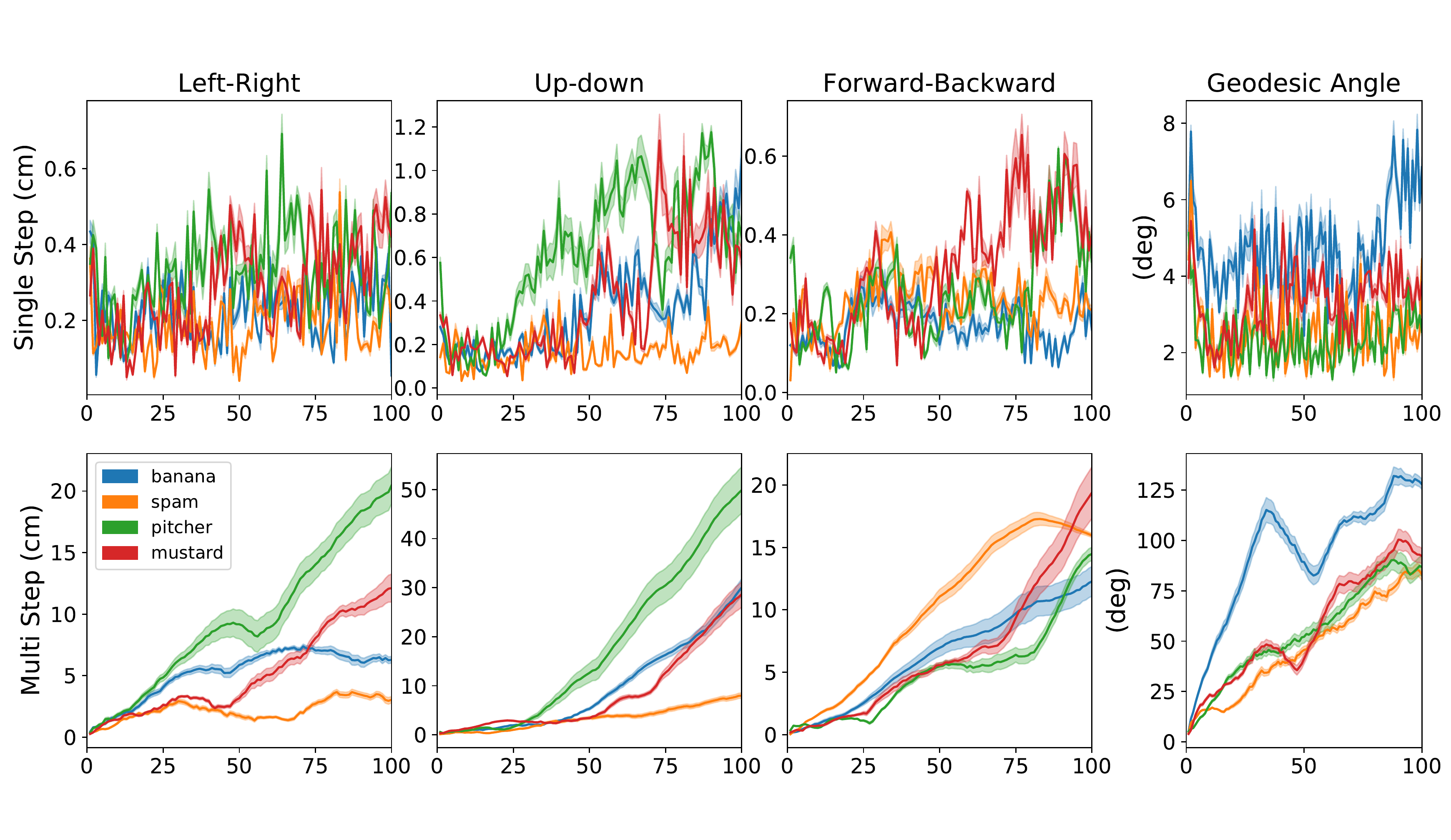}
        \vspace{-8mm}
        \caption[Real Robot Prediction Results]{Errors for single step and integrated (multi step) motion for the real robot setting without any fine-tuning, so the~\mnet~has never seen a robot during training. Four different objects (mentioned in the key) were all recorded completing four different 100 step motions including significant rotations and translations. The x axis correspods to the frame number in the sequence.} 
        \label{fig:real-pred}
        \vspace{-8mm}
    \end{center}
\end{figure*}

Figure~\ref{fig:real-pred} shows how much noisier results with real videos are because, as usual, there is far too little real data compared to the simulated data. Nevertheless, we see that unlike for the floating object where the left-right and up-down motion were essentially isotropic, for the real robot the errors are significantly worse in the up-down motion. We attribute this primarily to the fact that the lighting is significantly more affected by up-down motion than sideways (using ceiling-mounted lights). Since our domain randomization did not include any light changes either between or during sequences, it is not particularly surprising the~\mnets~are sensitive to lighting. Therefore, with a more sophisticated domain randomization system such as in~\cite{tremblay2018training}, these additional error sources can be mitigated. 

%When comparing the object, the pitcher was about twice the size of the mustard which in turn was about twice the size of the spam and banana (the last three of which are all YCB objects). It appears smaller, textured objects, like the spam, performed best, while the pitcher, which is completely blue, was more challenging. 

As is inevitable, there is a drop in performance when training on simulated data and testing on real data. However, since the networks never once saw a robot during training or any real image, these results show that the~\mnets~are able to track the 6D motion of arbitrary objects without any fine-tuning for around one second. Additionally, once again, the errors in the segmentation model are hidden within these translation and rotation errors, so with a more reliable segmentation model and more extensive domain randomization, performance could improve significantly.

\subsection{Control} \label{sec:ctrl_results}

\subsubsection{Floating-Object}

% \begin{figure}
%     \begin{center}
%         \includegraphics[width=1\textwidth]{figures/floating-control-examples-noocc-90.pdf}
%         \caption[Floating Object Control Examples]{These are six examples of the single object (no-occ) setting, with the letter in green when the sequence ended within the 5cm-$5^\circ$ criterion, and red otherwise.} 
%         \label{fig:float-ctrl-exp}
%     \end{center}
% \end{figure}

\begin{figure}
    \begin{center}
        \includegraphics[width=.36\textwidth]{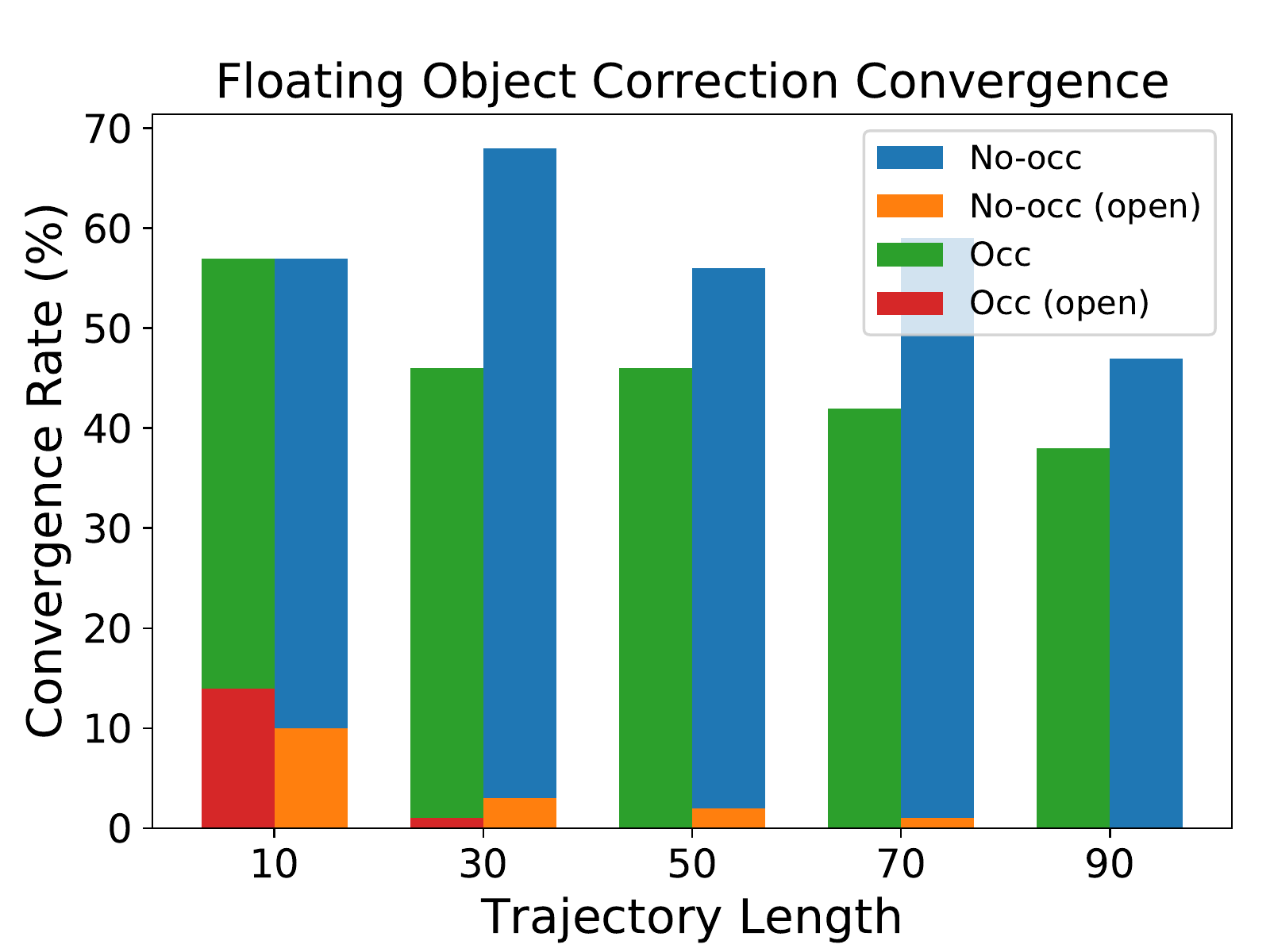}
        \vspace{-2mm}
        \caption[Floating Object Convergence Rates with Corrections]{The red and orange show how often the final pose was with 5cm and $5^\circ$ without any correction iterations. Meanwhile the green and blue show the convergence rate for the same success criterion after 100 correction iterations for the Occ (3 objects in the scene so distractors/occlusions present) and No-occ (single object in scene) setting.} 
        \label{fig:float-conv}
    \end{center}
\end{figure}

For the floating object, convergence was reached fairly reliably across all sequence lengths when using the iterative correction procedure (see figure~\ref{fig:float-conv}). Across the board, the objects in the multi object setting (Occ) converged after 100 correction iterations in 30-40\% of the sequences, while in the single object setting it converged 40-50\% of the time. This means even if the open-loop predictions get worse over time, for example tripling the perturbation length from 30 to 90, the correction iterations applied afterwards are still able to remove those additional errors.

The most common failure case by far is the geodesic angle error being upwards of $30^\circ$ or $40^\circ$ and then even though the corrections are able to recover the left-right and up-down position often to millimeter accuracy, the angle will never get fixed because the current and target simply look too different for the rotation model to make sense of the them. This is the crux of dealing with never-before-seen objects. The problem is that as long as an object is moving step by step~\mnets~can track the motion fairly well, but the model does not build up any kind of intuition for object continuity beyond rotations by a 5-$8^\circ$.

One promising way to prevent the correction procedure from getting stuck is to use uncertainty in our model. If the~\mnets~is particularly uncertain about what motion occurred between the current and target frames, then the a more aggressive search procedure can be used to try rotating the object in different directions until a match is found. While such a search procedure can be more difficult, it can be informed by the model's uncertainty in the motion of the object so far, and it will prevent the system from getting stuck.

\subsubsection{Simulated PR2}

\begin{figure}
    \begin{center}
        \includegraphics[width=.36\textwidth]{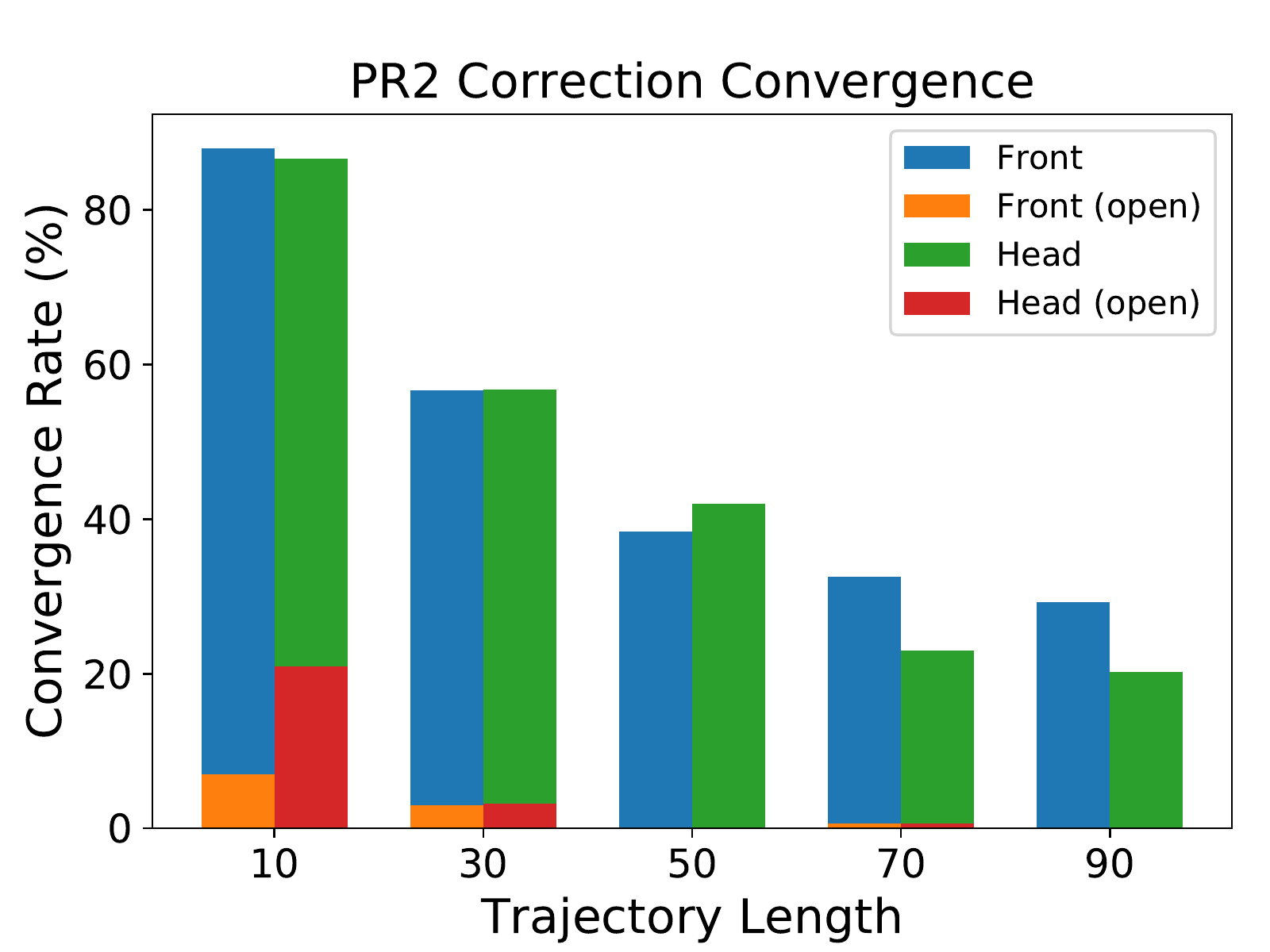}
        \vspace{-2mm}
        \caption[Simulated PR2 Correction Convergence]{The red and orange show how often the final pose was with 5cm and $5^\circ$ without any correction iterations. Meanwhile the green and blue show the convergence rate for the same success criterion after 20 correction iterations for the Front (third-person view of the robot) and Head (camera mounted on the robots head) views.} 
        \label{fig:pr2-conv}
    \end{center}
\end{figure}

Running control on the PR2 exhibited many of the same issues as with the floating object in the occluded setting. However, now that the motion of the distractor objects (eg. robot arm) is highly correlated with the object motion, the predicted segmentation degrades faster, which compounds the rotation and translation errors. In large part due to the errors in the segmentation model, the convergence rate drops significantly as the perturbation length is increased (see figure~\ref{fig:pr2-conv}). Interestingly there doesn't seem to be a major difference in performance between the first and third person views of the robot. Just as with the floating object setting, the correction procedure improves the performance significantly. We especially noticed that even though the open-loop predictions were not particularly accurate in forward-backward motion (since there is no depth information), the correction procedure is consistently able to improve the position of the object in all three dimensions, and that if the correction fails, it is almost always because of the error in the orientation.

% \begin{figure}
%     \begin{center}
%         \includegraphics[width=1\textwidth]{figures/pr2-control-examples-front-90.pdf}
%         \caption[Simulated PR2 Front View Examples]{These are six examples of the simulated PR2 in its front (third person) view, with the letter in green when the sequence ended within the 5cm-$5^\circ$ criterion, and red otherwise.} 
%         \label{fig:front-exm}
%     \end{center}
% \end{figure}

% \begin{figure}
%     \begin{center}
%         \includegraphics[width=1\textwidth]{figures/pr2-control-examples-head-90.pdf}
%         \caption[Simulated PR2 Head View Examples]{These are six examples of the simulated PR2 in its head (first person) view, with the letter in green when the sequence ended within the 5cm-$5^\circ$ criterion, and red otherwise.} 
%         \label{fig:head-exm}
%     \end{center}
% \end{figure}

\section{Conclusion}

While the predictions of~\mnets~are not nearly as accurate as other pose estimation methods such as~\cite{li2018deepim,byravan2017se3}, they are able to get noisy estimates from only RGB, and without the 3D model of the object being tracked. This enables~\mnets~to be applicable for many more settings, such as for an active learning task, where a new object is introduced and the robot must interact with it. Here the segmentation and motion estimates can provide a good first approximation of the 6D motion of the object, allowing the robot to build an intuition of the shape and motion of the novel object.

However, we do believe that there is ample room for improvement, before~\mnets~can compete with other methods that require additional information about the object or environment. One of the biggest issues is that the single step motion estimates are open-loop, so the errors in the estimated pose of the object accumulate over time. We hope to close the loop using feedback in a fashion similar to the iterative correction procedure; we are currently investigating this possibility. To improve multi-step predictions we can also try using some uncertainty information, to allow the model to focus on particularly challenging parts of the video and possibly revise the predicted motion. Additionally, our data generation process can be improved to include changes in lighting and more dynamic backgrounds, to improve the transfer capabilities.

% the initial mask can be relaxed to be a bounding box

% a promising application is an online setting, where an objects position must only be tracked for a few seconds, such as for grasping - after the object has been grasped, the pose can be inferred from the robot kinematics
% mnets also provide a good first approximation of the motion of unknown objects, which is a good stepping stone for open-ended learning settings where new objects can appear and it can take a few experiences before the robot can recognize
% avoid many symmetry issues since single step motions are usually small
% not many baselines available

\section*{Acknowledgment}

This work was funded in part by the National Science
Foundation under contract number NSF-NRI-1637479 and
STTR number 1622958 awarded to LULA robotics.

\bibliographystyle{./bibliography/IEEEtran}
\bibliography{./bibliography/IEEEabrv,./bibliography/IEEEexample}

% \vspace{12pt}
% \color{red}
% IEEE conference templates contain guidance text for composing and formatting conference papers. Please ensure that all template text is removed from your conference paper prior to submission to the conference. Failure to remove the template text from your paper may result in your paper not being published.

\onecolumn

\section*{Appendix}
To give the reader a better idea of what our tasks and experiments looked like, we provide the following examples.

% We additionally include a figure of how our iterative correction scheme helps the pose errors decrease.

% \begin{figure*}[b]
% \centering
% \begin{minipage}{.5\textwidth}
%   \centering
%   \includegraphics[width=.8\linewidth]{figures/floating-control-noocc-correction-errors.pdf}
%   \captionof{}{Unoccluded (single object)}
%   \label{fig:test1}
% \end{minipage}%
% \begin{minipage}{.5\textwidth}
%   \centering
%   \includegraphics[width=.8\linewidth]{figures/floating-control-occ-correction-errors.pdf}
%   \captionof{}{Occluded (multi-object)}
%   \label{fig:test2}
% \end{minipage}
% \caption[Errors of Correction Iterations of the Floating Object]{These figures show how the error between the original target pose and the current pose decreases as the corrections are iterated and for different perturbation lengths (in the key). The red dashed line shows the 5cm-$5^\circ$ criterion for convergence. Note that here only the runs that did end up converging were averaged.} % , the failed runs are not included in the plot at al\textbf{}
% \label{fig:float-cor-err}
% \end{figure*}

\begin{figure*}
    \begin{center}
        \includegraphics[width=1\textwidth]{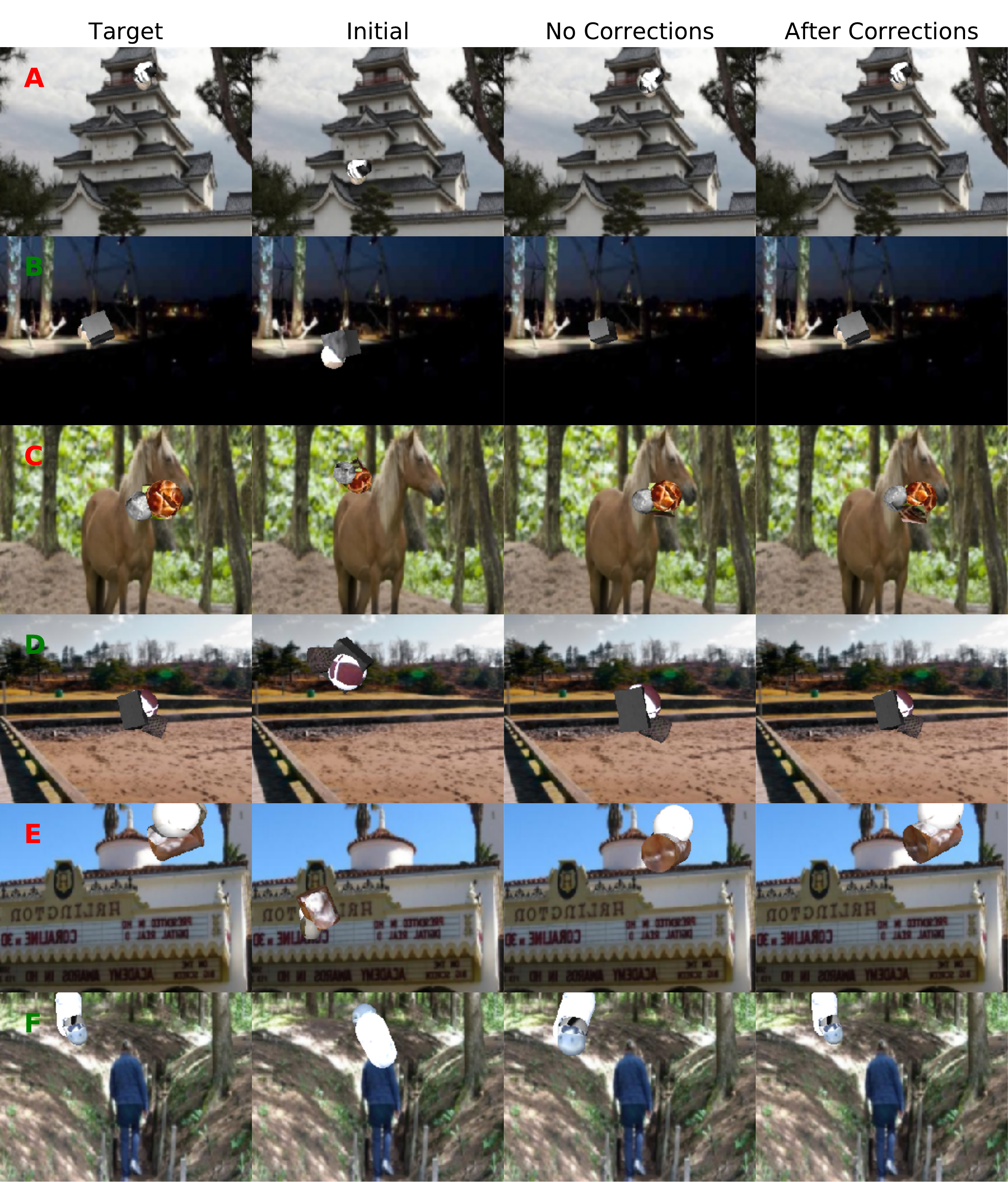}
        \caption[Floating Object Control Examples]{These are six examples of the single floating object (no-occ) setting, with the letter in green when the sequence ended within the 5cm-$5^\circ$ criterion, and red otherwise.} 
        \label{fig:float-ctrl-exp}
    \end{center}
\end{figure*}

\begin{figure*}
    \begin{center}
        \includegraphics[width=1\textwidth]{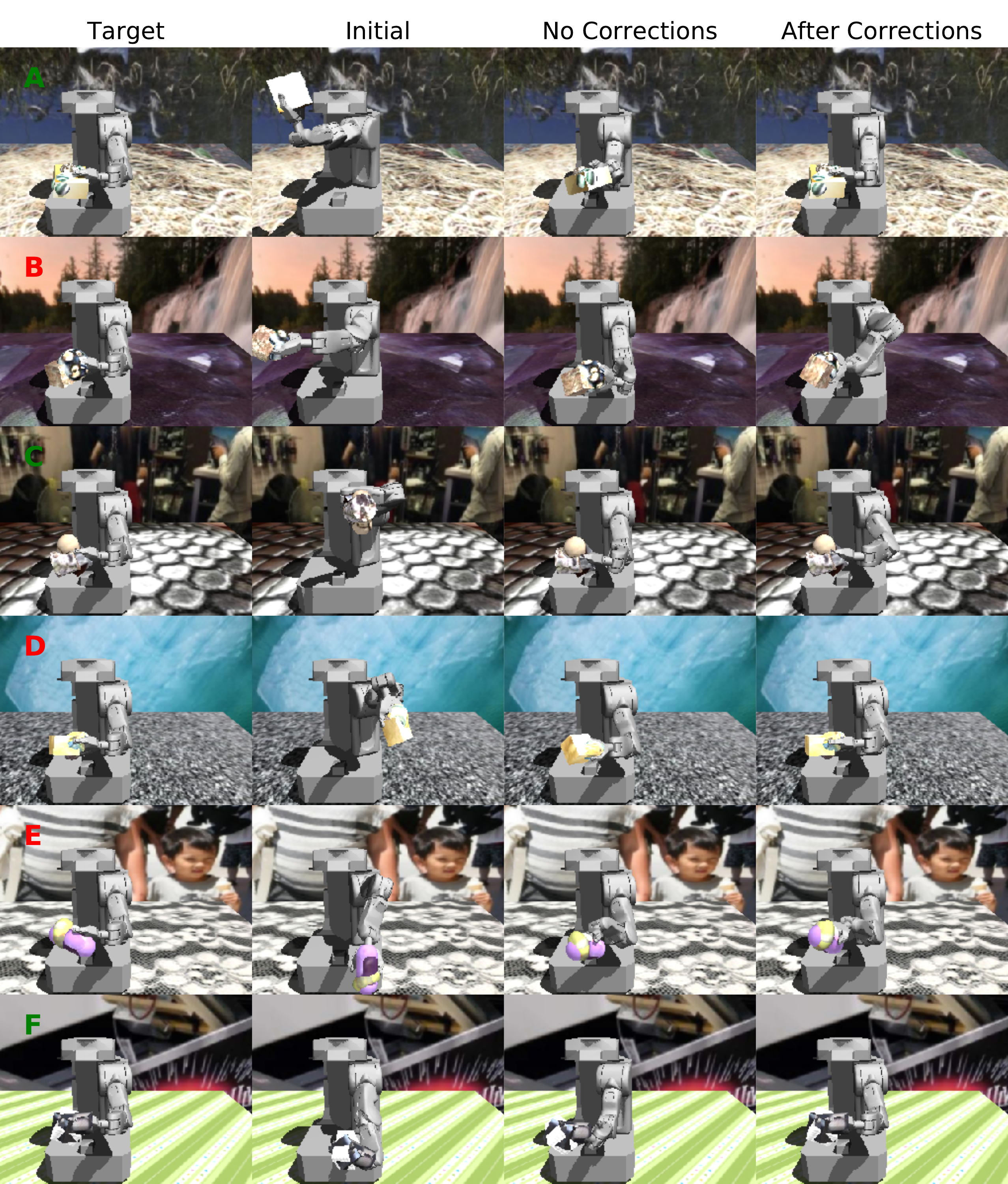}
        \caption[Simulated PR2 Front View Examples]{These are six examples of the simulated PR2 in its front (third person) view, with the letter in green when the sequence ended within the 5cm-$5^\circ$ criterion, and red otherwise.} 
        \label{fig:front-exm}
    \end{center}
\end{figure*}

\begin{figure*}
    \begin{center}
        \includegraphics[width=1\textwidth]{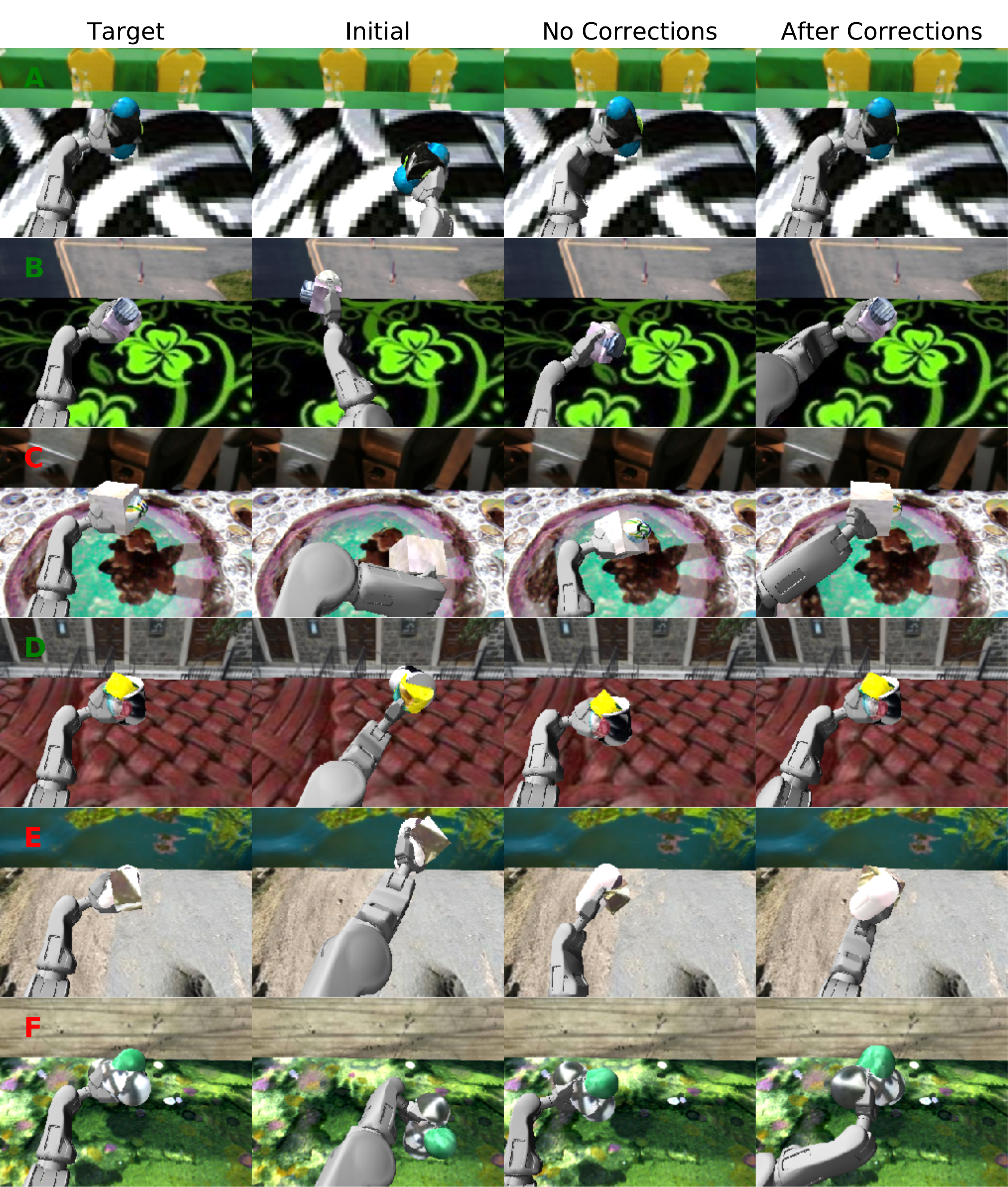}
        \caption[Simulated PR2 Head View Examples]{These are six examples of the simulated PR2 in its head (first person) view, with the letter in green when the sequence ended within the 5cm-$5^\circ$ criterion, and red otherwise.} 
        \label{fig:head-exm}
    \end{center}
\end{figure*}

\end{document}